\documentclass[runningheads]{llncs}

\usepackage{eccv}

\usepackage{eccvabbrv}
\usepackage{amsmath}
\usepackage{amssymb}
\usepackage{booktabs}
\usepackage{pifont}
\usepackage{algorithm}
\usepackage{algpseudocode}
\usepackage{marvosym}
\usepackage{multirow}
\usepackage{stfloats}
\usepackage{caption}
\usepackage{soul}
\usepackage{colortbl}

\usepackage{graphicx}
\usepackage{booktabs}

\usepackage{marvosym}
\usepackage{eccvabbrv}
\usepackage{multirow}
\usepackage{graphicx}
\usepackage{booktabs}

\usepackage[accsupp]{axessibility} 
\usepackage{algorithm}
\usepackage{algpseudocode}
\usepackage{wasysym}
\usepackage{enumitem}
\usepackage{amsmath}
\usepackage{bm}
\usepackage{microtype} 

\usepackage[pagebackref, breaklinks, colorlinks, citecolor = eccvblue]{hyperref}
\usepackage{orcidlink}
\usepackage{wrapfig}

\begin{document}

\title{Zero-shot Object Counting with Good Exemplars} 



\author{Huilin~Zhu\inst{1,2,3,\dag}\orcidlink{0000-0003-1607-7283} \and
Jingling~Yuan\inst{1,2,\dag}\orcidlink{0000-0001-7924-8620} \and
Zhengwei~Yang\inst{4,\dag}\orcidlink{0000-0002-8190-1438} \and
Yu~Guo\inst{3,5}\orcidlink{0000-0002-0642-7684} \and
Zheng~Wang\inst{4}\orcidlink{0000-0003-3846-9157} \and
Xian~Zhong\inst{1,2,6(\textrm{\Letter})}\orcidlink{0000-0002-5242-0467} \and
Shengfeng~He\inst{3(\textrm{\Letter})}\orcidlink{0000-0002-3802-4644}
}


\authorrunning{H.~Zhu \etal}



\institute{
Sanya Science and Education Innovation Park, Wuhan University of Technology \and
Hubei Key Laboratory of Transportation Internet of Things, School of Computer Science and Artificial Intelligence, Wuhan University of Technology
\email{zhongx@whut.edu.cn} \and
School of Computing and Information Systems, Singapore Management University 
\email{shengfenghe@smu.edu.sg} \and
School of Computer Science, Wuhan University \and
School of Navigation, Wuhan University of Technology \and
ROSE@EEE, Nanyang Technological University \\
$^\dag$ Equal Contribution \\
\url{https://github.com/HopooLinZ/VA-Count}}
\maketitle

\begin{abstract}
Zero-shot object counting (ZOC) aims to enumerate objects in images using only the names of object classes during testing, without the need for manual annotations. However, a critical challenge in current ZOC methods lies in their inability to identify high-quality exemplars effectively. This deficiency hampers scalability across diverse classes and undermines the development of strong visual associations between the identified classes and image content. To this end, we propose the Visual Association-based Zero-shot Object Counting (VA-Count) framework. VA-Count consists of an Exemplar Enhancement Module (EEM) and a Noise Suppression Module (NSM) that synergistically refine the process of class exemplar identification while minimizing the consequences of incorrect object identification. The EEM utilizes advanced vision-language pretaining models to discover potential exemplars, ensuring the framework's adaptability to various classes. Meanwhile, the NSM employs contrastive learning to differentiate between optimal and suboptimal exemplar pairs, reducing the negative effects of erroneous exemplars. 
VA-Count demonstrates its effectiveness and scalability in zero-shot contexts with superior performance on two object counting datasets.
\end{abstract}

\section{Introduction}
In visual monitoring applications, object counting plays a critical role in analyzing images or videos. Traditional methods focus on high precision within predefined object categories, such as crowds, vehicles, and cells~\cite{tyagi2023degpr, arteta2016counting, babu2022completely, tian2023knowledge, xiong2024glance}. Yet, these methods are limited to specific categories, lacking the flexibility to adapt to new, unseen classes.
To address these challenges, class-agnostic methods have been developed for scenarios with unseen classes. These methods, including few-shot, reference-free, and zero-shot object counting~\cite{ranjan2022exemplar, shi2022represent, yang2021class, you2023few, huang2023point}, provide varying levels of independence from predefined object classes.

\begin{figure}[t]
	\centering
	\includegraphics[width =0.95 \textwidth]{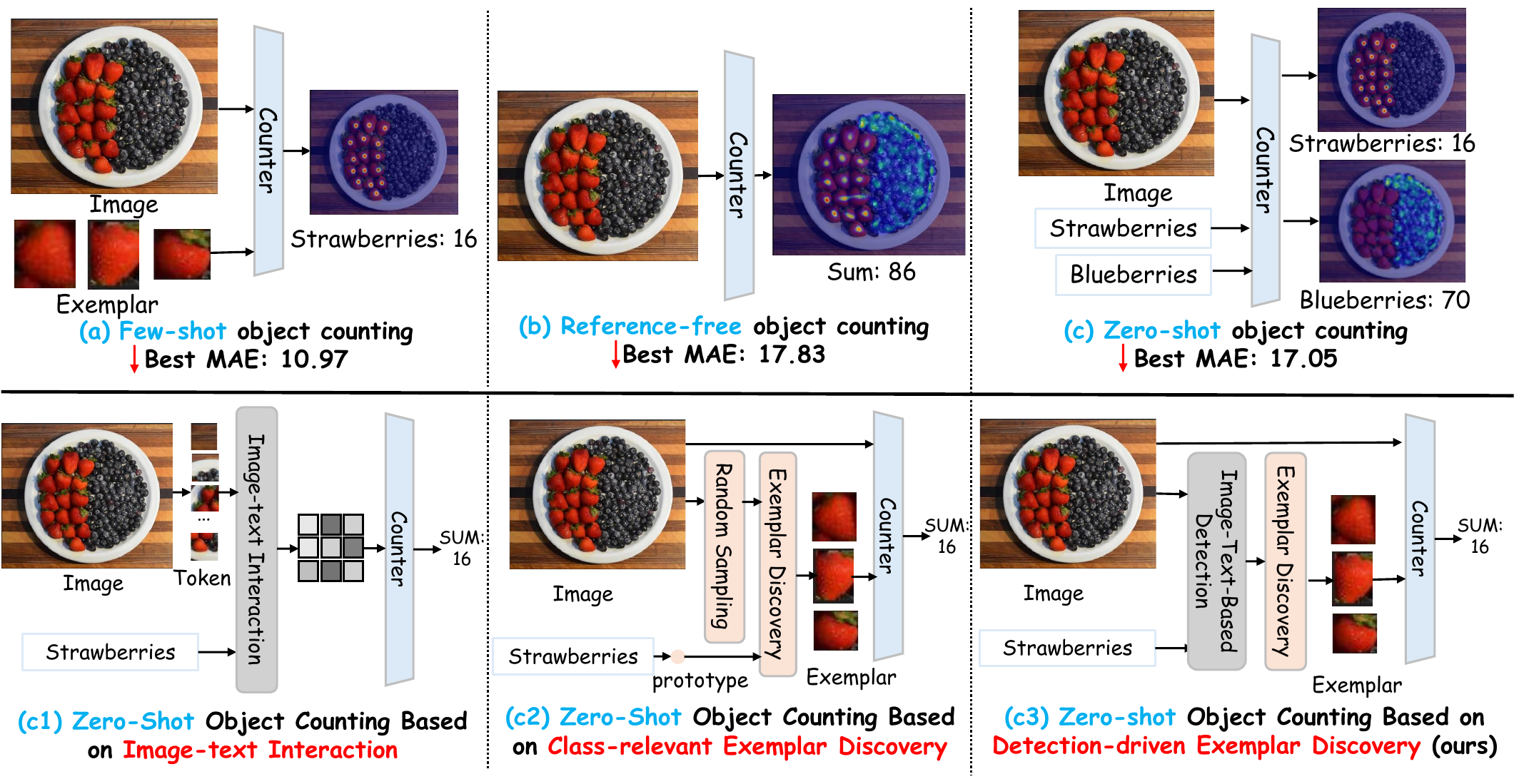}
	\caption{Illustration of class-agnostic object counting methods. (a) Few-shot uses limited annotations for counting. (b) Reference-free quantifies objects without annotations. (c) Zero-shot counts specific classes without annotations, further divided into: (c1) Image-text association, leveraging direct image-text correlations. (c2) Class-related exemplar search, using prototypes to link classes with images. (c3) Our method introduces a detection-driven exemplar discovery to harmonize text with visual representations, distinguishing it from prior methods.}
	\label{fig:1}
\end{figure}

In this context, different strategies are adopted for object counting under varying constraints, as illustrated in \cref{fig:1}. Few-shot counting methods~\cite{yang2021class, nguyen2022few, you2023few}, depicted in \cref{fig:1}(a), method the task as a matching problem, using a small number of annotated bounding boxes to identify and count objects throughout the image. While effective, this method requires fine-tuning with annotations from novel classes, limiting its scalability in real-world surveillance settings due to the sparse availability of annotated bounding boxes.
To circumvent the limitations of bounding box annotations, reference-free counting methods are developed~\cite{ranjan2022exemplar, hobley2022learning, liu2022countr, djukic2023low}, as shown in \cref{fig:1}(b). These methods aim to ascertain the total number of objects in an image without relying on specific cues. Nevertheless, the lack of specificity in counting categories makes these methods prone to errors induced by background noise, as they indiscriminately count all visible objects, leading to a lack of control in the counting process.

In pursuit of more scalable and realistic counting solutions, zero-shot methods~\cite{Bansal18zero, zheng2021zero}, illustrated in \cref{fig:1}(c), are introduced. These techniques are designed to count objects from specified classes within an image without prior annotations for those classes, addressing the limitations of both few-shot and reference-free methods by providing enhanced specificity and scalability.
These methods can be categorized into two streams. The initial method~\cite{jiang2023clip, kang2023vlcounter} leans on image-text alignment to comprehend object-related correlations without needing physical exemplars. This method enhances scalability for unidentified classes but struggles with adequately representing image details for target classes, especially those with atypical shapes, as demonstrated in \cref{fig:1}(c1). Conversely, the second method~\cite{xu2023zero} concentrates on identifying objects through the discovery of class-relevant exemplars. This is achieved by creating pseudo labels that assess the resemblance between image patches and class-generated prototypes. Nevertheless, this method's reliance on arbitrary patch selection hampers its ability to accurately outline entire objects. Additionally, the absence of direct text-image engagement restricts its scalability, tethered to the pre-defined categories present in the training dataset, as illustrated in \cref{fig:1}(c2).

As shown in \cref{fig:1}(c3), we introduce the Visual Association-based Zero-shot Object Counting (VA-Count) framework. VA-Count aims to create a robust link between specific object categories and their corresponding visual representations, ensuring adaptability to various classes. This framework is anchored by three core principles. First, it prioritizes flexibility and scalability, enabling adaptation to novel classes beyond its initial parameters. Second, it enhances precision in identifying exemplary objects, strengthening the connection between visual depictions and their categories. Third, it devises strategies to reduce the effects of localization errors on counting precision.
Building on these principles, VA-Count integrates an Exemplar Enhancement Module (EEM) and a Noise Suppression Module (NSM), which are dedicated to refining exemplar identification and mitigating adverse impacts, respectively.

In detail, the EEM expands VA-Count's capacity to handle various classes through the integration of Vision-Language Pretaining (VLP) models, such as Grounding DINO~\cite{Liu2023DINO}. These VLP models, trained on extensive datasets, excel in identifying a wide range of classes by defining specific categories. In the context of ZOC, it is essential to select exemplars that each contain precisely one object from among the potential bounding boxes that might encompass varying object quantities. To this end, we deploy a binary filter aimed at rigorously refining the set of candidate exemplars, excluding those that fail to comply with the single-object requirement. This filtration step is pivotal for ensuring the precision and consistency necessary for ZOC.

Moreover, even when potential exemplars accurately represent single objects, the unintentional inclusion of exemplars not pertaining to the target category poses a persistent problem. This misalignment introduces uncertainty into the learning process that associates exemplars with images. To counteract this issue, the NSM module operates as a safeguard by identifying negative exemplars, which are unrelated to the intended category. Contrasting with the EEM, which focuses on selecting ideal samples to foster visual connections with images, the NSM employs samples from irrelevant classes to build these associations, utilizing contrastive learning to differentiate between them. This method of contrastive learning acts as a rectifying mechanism, markedly improving the accuracy and efficiency of the associative learning framework.

In summary, our contributions are threefold:
\begin{itemize}
\item We introduce a Visual Association-based Zero-shot Object Counting framework, which facilitates high-quality exemplar identification for any class without needing annotated examples and forges robust visual connections between objects and images.
\item We propose an exemplar enhancement model leveraging the universal class-agnostic detection capabilities of the Vision-Language Pretaining model for precise exemplar selection, and a Noise Suppression Module to minimize the adverse effects of incorrect samples in visual associative learning.
\item Extensive experiments conducted on two object counting datasets demonstrate the state-of-the-art accuracy and generalizability of VA-Count, underscoring its notable scalability.
\end{itemize}

\section{Related Work}
\subsection{Class-Specific Object Counting}

Object counting plays a crucial role in public safety, public administration, and the liberation of human labor. 
Currently, class-specific object counting~\cite{ranjan2022exemplar, shi2022represent, yang2021class, you2023few,liu2022reducing} is the predominant method, which entails identifying specific object categories (such as humans~\cite{liu2020adaptive, zhu2023daot, ranjan2018iterative, zhu2023find, liu2023fine}, vehicles~\cite{zhang2016vision, mundhenk2016large}, fishes~\cite{Sun_2023_CVPR}, cells~\cite{tyagi2023degpr}, \etc.) leveraging object detection or density estimation and counting accordingly. While these methods show excellence within close-set scenarios with a fixed number of categories, transferring them to arbitrary categories poses challenges.
Introducing novel categories necessitates retraining or fine-tuning a counting model with new data, which limits their applicability in real scenarios.

\subsection{Class-Agnostic Object Counting}

Class-agnostic object counting~\cite{lu2019class, gong2022class, nguyen2022few,shi2024training,wang2024vision} is proposed for scenarios with less data, which can be divided into few-shot and zero-shot depending on the annotation usage. Specifically, GMN~\cite{lu2019class} initially frames the class-agnostic counting task as a matching task, leading to FamNet~\cite{ranjan2021learning}, which implements ROI Pooling for broad applicability across \textsc{FSC-147}. As multi-class datasets emerged, the focus shifts towards few-shot methods, where LOCA~\cite{djukic2023low} enhances feature representation and exemplar adaptation; and CounTR~\cite{liu2022countr} utilizes transformers for scalable counting with a two-stage training model.
BMNet~\cite{shi2022represent} innovates with a bilinear matching network for refined object similarity assessments. In the realm of zero-shot methods, which are categorized into two types, methods like ZSC~\cite{xu2023zero} leverage textual inputs to generate prototypes and filter image patches, thus reducing the need for extensive labeling, albeit with fixed generators that limit scalability. CLIP-Count~\cite{jiang2023clip} employs CLIP to encode text and images separately, establishing semantic associations crucial for intuitive counting. VLCount~\cite{kang2023vlcounter} takes this further by enhancing CLIP's text-image association learning specifically for object counting. Additionally, PseCo~\cite{huang2023point} introduces a SAM-based multi-task framework that achieves segmentation, dot mapping, and detection on counting data, offering broad application prospects but also necessitating greater computational resources.

\subsection{Vision-Language Pretaining Model}
In recent years, Vision-Language Pretaining (VLP) methods have proven pivotal in enhancing scene understanding and representation learning capabilities. Their adaptability makes them applicable across a wide range of downstream tasks~\cite{ming2022delving, du2022learning, song2022vision, dou2022coarse, li2023clip, he2023region, chen2023towards, bai2023relation, xie2023unified}. CLIP~\cite{Rad2021CLIP} segregates vision and language features, aligning them through contrastive learning. BLIP~\cite{Li2022BLIP} introduces a multimodal mixture of encoders and decoders to align different modalities. Building upon this, BLIP2~\cite{Li2023BLIP2} combines specialized vision and language models to enhance multimodal understanding capabilities through bootstrapping. Grounding DINO~\cite{Liu2023DINO} incorporates language into close-set detection, improving generalization for open-set detection. 
The Segment Anything Model (SAM)~\cite{Alex2023segment} is based on a prompt-based segmentation task, allowing flexible prompts for zero-shot capabilities across diverse tasks. VLP models, known for their robust multimodal comprehension and scene understanding, significantly advance deep learning and facilitate learning of unknown classes.

\begin{figure}[t]
	\centering
	\includegraphics[width = 0.96\textwidth]{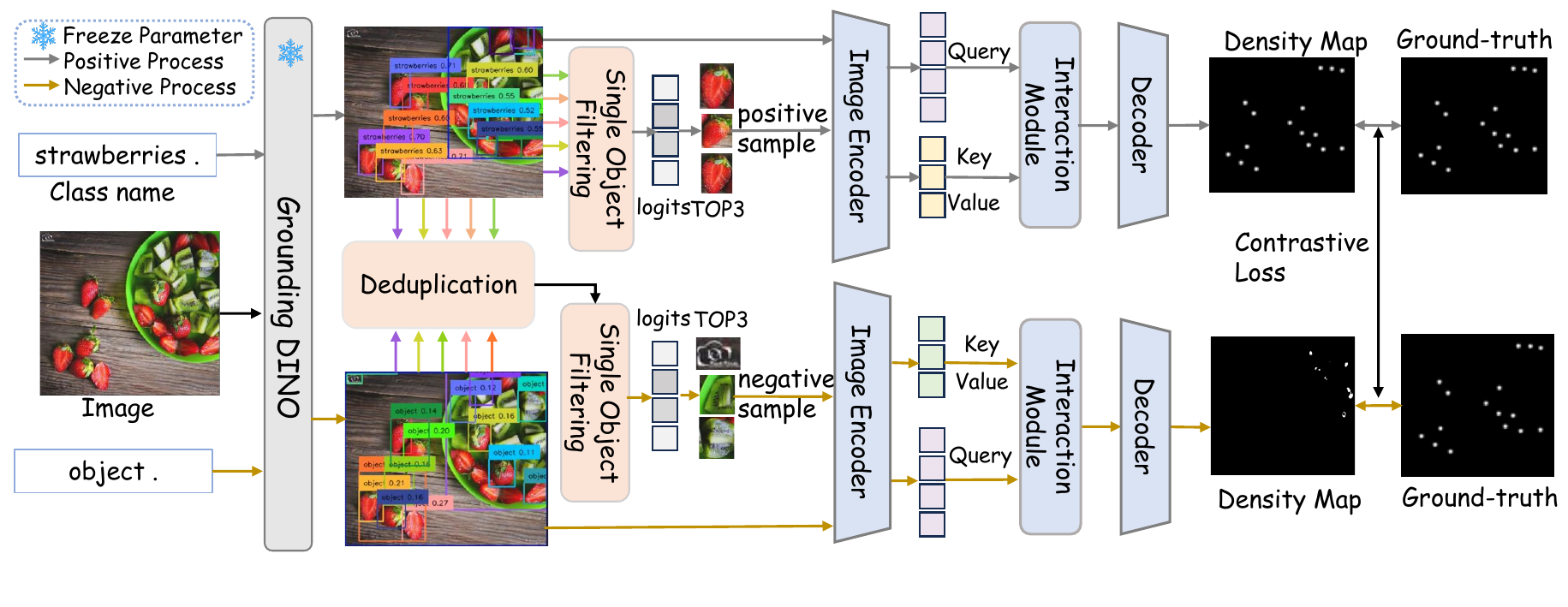} 
	\caption{Overview of the proposed method. The proposed method focuses on two main elements: the Exemplar Enhancement Module (EEM) for improving exemplar quality through a patch selection integrated with Grounding DINO~\cite{Liu2023DINO}, and the Noise Suppression Module (NSM) that distinguishes between positive and negative class samples using density maps. It employs a Contrastive Loss function to refine the precision in identifying target class objects from others in an image.}
	\label{fig:Framework}
\end{figure}

\section{Proposed Method}
\subsection{Formula Definition}
\begin{algorithm}[t]
\caption{Grounding DINO-Guided Exemplar Enhancement Module}
\small
\begin{algorithmic}[1]
\State $I$: Input image
\State $T^p$: Positive text label (\{specific class\}), $T^n$: Negative text label (``object'')
\State $B^p$: Bounding boxes for positive samples, $\mathcal{S}^p$: Logits for positive samples
\State $B^n$: Bounding boxes for negative samples, $\mathcal{S}^n$: Logits for negative samples
\State $\tau_l$: Logits threshold, $\tau_\mathrm{iou}$: IoU threshold
\State $M(\cdot)$: Single Object Classifier
\State \textbf{Input:} $I$, $T^p$, $T^n$
\State \textbf{Output:} $\mathcal{O}^p = \{(B^p, \mathcal{S}^p)\}$: Positive outputs, $\mathcal{O}^n = \{(B^n, \mathcal{S}^n)\}$: Negative outputs

\State \textbf{Grounding DINO Process:}
\State $F \gets \mathrm{ExtractFeatures}(I)$
\State $\mathcal{S}^p, B^p \gets \mathrm{Detect}(F, T^p)$, filter by $\tau_l; and~\mathcal{S}^n, B^n \gets \mathrm{Detect}(F, T^n)$, filter by $\tau_l$
\State \textbf{Deduplication and Filtering:}
\State Initialize $B^n_\mathrm{filtered}, B^p_\mathrm{new}, B^n_\mathrm{new}$
\For{$b^n$ in $B^n$} \Comment{Remove duplicates}
 \If{$b^n$ is unique in $B^p$ with IoU $< \tau_\mathrm{iou}$}
 \State $B^n_\mathrm{filtered}$.append($b^n$)
 \EndIf
\EndFor

\ForAll{$b \in B^p \cup B^n_\mathrm{filtered}$} \Comment{Single object filter}
 \If{$M(b)$ is true} \State Add $b$ to the appropriate new set
 \EndIf
\EndFor

\State Update $\mathcal{O}^p, \mathcal{O}^n$ with new sets
\end{algorithmic}
\label{alg}
\end{algorithm}
As shown in \cref{fig:Framework}, we introduce a Visual Association-based Zero-shot Object Counting framework (VA-Count) focusing on zero-shot, class-agnostic object counting. The categories among the training set $C_\mathrm{train}$, validation set $C_{\mathrm{val}}$, and testing set $C_{\mathrm{test}}$ are distinguished, ensuring no overlap among them $(C_\mathrm{train} \cap C_{\mathrm{val}} \cap C_{\mathrm{test}} = \emptyset)$. VA-Count generates density maps $D$ from input images $I$ for any given class $C$, and counts objects using these density maps.
Specifically, VA-Count utilizes pseudo-exemplars $E^p$ to enhance image-text associations, acting as a bridge to establish robust visual correlations between $E^p$ and the images $I$. 
To extract exemplars from images, we propose the use of two key modules: the Exemplar Enhancement Module (EEM) (\cf \cref{Sec3.3}) and the Noise Suppression Module (NSM) (\cf \cref{Sec3.4}).

To alleviate the noise introduced by objects belonging to other classes on the target objects within images, the EEM and NSM are simultaneously used to obtain positive exemplars $B^p$ and negative exemplars $B^p$.
The EEM consists of Grounding DINO $G(\cdot)$ and a filtering module $\Phi(\cdot)$. There are different filtering modules for positive and negative samples $\Phi^p(\cdot)$ and $\Phi^n(\cdot)$ respectively. $\Phi^p(\cdot)$ is a binary classifier, while $\Phi^n(\cdot)$ consists of a binary classifier and a deduplication module. The two kinds of pseudo-exemplars and images are then fed into the Counter $\Gamma(\cdot)$ simultaneously for correlation learning. $\Gamma(\cdot)$ comprises an image encoder, correlation module, and decoder. 
The optimization goal of this paper is as follows, where $\mu(\cdot)$ denotes the similarity, and $D^p, D^n, D^g$ represent the density maps for positive, negative, and ground truth respectively:
\begin{equation}
 	D^p = \Gamma \left(\Phi^p \left(G \left(I, T^p \right) \right) \right), \quad
 	D^n = \Gamma \left(\Phi^n \left(G \left(I, T^n \right) \right) \right), 
\end{equation}
\begin{equation}
 	\mathrm{Objective} = 
 	\begin{cases}
 	\max \mu(D^p, D^g), \\
 	\min \mu(D^n, D^g).
 	\end{cases}
\end{equation}



\subsection{Exemplar Enhancement Module}
\label{Sec3.3}



We introduce an Exemplar Enhancement Module (EEM) for detecting objects within images and refining the detected objects as target exemplars. The workflow of the EEM is outlined in \cref{alg}. The EEM ensures VA-Count's scalability to arbitrary classes by incorporating Vision-Language Pretaining (VLP) models (\eg, Grounding DINO~\cite{Liu2023DINO}) for potential exemplar discovery, renowned for its efficiency in feature extraction and precision in object localization. Furthermore, the EEM involves meticulously discovering and refining potential exemplars to enhance the quality of positive and negative exemplars for precise object counting.

\textbf{Grounding DINO-Guided Box Selection.}
Given the training set input image $I_i$, accompanied by predefined sets of positive text labels $T_i^p = \{C_i\}$ and negative text labels $T_i^n = {\mathrm{``object''}}$, where $C_i$ represents the specified target class for the input image and $T_i^n$ is fixed as ``object''. These labels correspond to the target objects and the noise objects, respectively. 
Taking positive exemplar discovery as an example, Grounding DINO assigns logits value $\mathcal{S}^p_i = \{s_{i, j}\}_{j = 0}^{m}$ to all candidate bounding boxes $B^p_i = \{b_{i, j}\}_{j = 0}^{m}$ based on $T_i^p$, $m$ denotes the number of candidate boxes within the image. For the $j$-th box in the $i$-th image, $s_{i, j}$ represents the likelihood that $b_{i, j}$ belongs to the specified class text $C_i$. The output of positive candidate boxes $\mathcal{O}^p$ can be formulated as:
\begin{equation}
\label{eq:positive_patch}
\mathcal{O}^p = \{G(I_i, T^p_i)\}_{i = 0}^k = \{{(B^p_i, \mathcal{S}^p_i)}\}_{i = 0}^k, 
\end{equation}
where $k$ denotes the number of images in the training set.

\textbf{Negative Samples and Deduplication.}
To minimize the impact of irrelevant classes on the counting accuracy of the target object, we adopt a filtering method for negative samples. Initially, we obtain all candidate bounding boxes for objects within each image. Similar to \cref{eq:positive_patch}, the negative candidate boxes $\mathcal{O}^n$ without filtering can be formulated as:
\begin{equation}
\label{eq5}
 \mathcal{O}^n = \left\{G \left(I_i, T^n_i \right) \right\}_{i = 0}^k = \left\{\left(B^n_i, \mathcal{S}^n_i \right) \right\}_{i = 0}^k, 
\end{equation}
where for each image $I_i$, the term $T_i^n$ = ``object'' is employed to identify and generate all bounding boxes $B^n$ within that image. This method guarantees the detection of bounding boxes for all objects present in the image.

Then, for each image $I_i$, we assess each bounding box $b^n$ from the negative candidate boxes $B^n$, and each $b^n$ is evaluated to determine its uniqueness in relation to the boxes within $B^p$. 
Specifically, a bounding box is deemed unique if its overlap with any box in $B^p$ is minimal, based on the Intersection over Union (IoU) threshold $\tau_\mathrm{iou}$, which can be formulated as:
\begin{equation}
 \mathrm{IoU} \left(B^p, B^n \right) = \frac{B^p \cap B^n}{B^p \cup B^n}, 
\end{equation}
where $B^p \cap B^n$ and $B^p \cup B^n$ denotes the intersection and union between positive $ B^p $ and negative $ B^n $ boxes. Unique negative boxes $ b^n $ are then included in the final set $ B^n_\mathrm{filtered} $ of negative exemplars.


\textbf{Single Object Exemplar Filtering.}
While DINO excels at identifying targets for arbitrary classes, each candidate box does not always contain a single object because boxes encompassing multiple objects may carry higher confidence levels than boxes of single objects. To ensure the integrity of the visual connections established with images, it's imperative to select exemplars that exclusively contain a single object. 
To achieve this, we treat singular discrimination as a binary classification task, using the binary classifier $\delta(\cdot)$ to refine candidate bounding boxes, ensuring each exemplar contains a single object.

\begin{wrapfigure}{r}{0.5\textwidth}
	\centering
	\includegraphics[width = 0.46\textwidth]{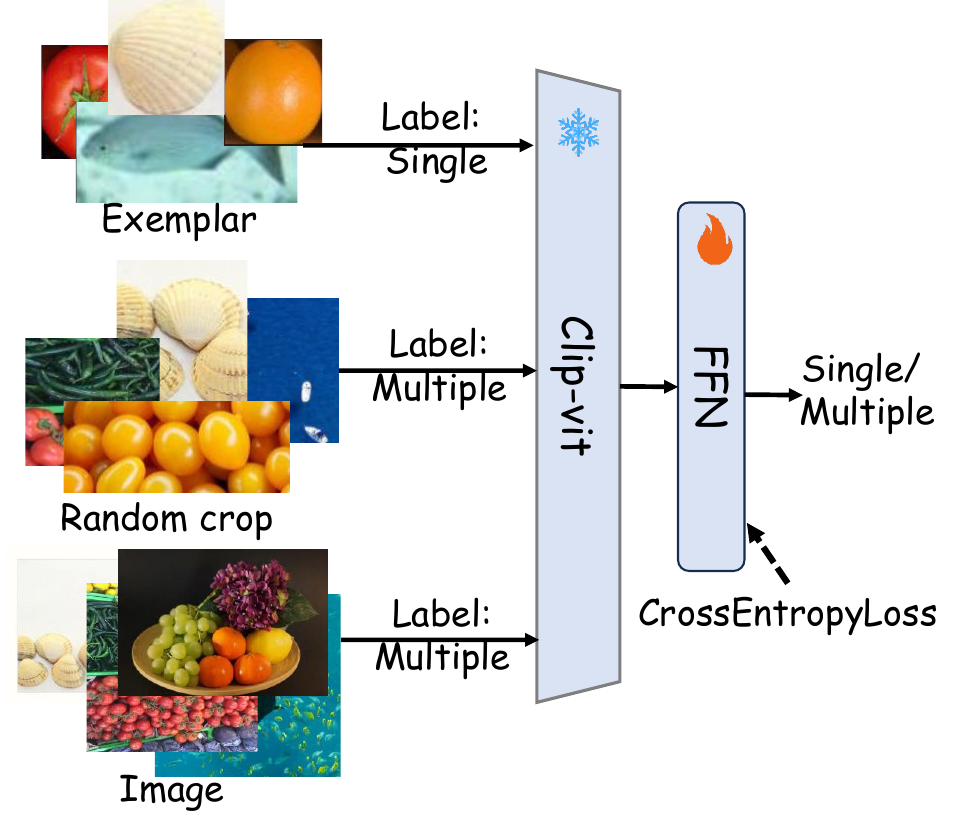}
	\caption{Illustration of the single object exemplar filtering with a frozen Clip-vit encoder and a trainable FFN to distinguish single from multiple objects.}
	\label{fig:Fra}
\end{wrapfigure}

As shown in \cref{fig:Fra}, $\delta(\cdot)$ leverages a frozen $\mathrm{Clip}\text{-}\mathrm{vit}$ backbone, integrated with a trainable Feed-Forward Network (FFN) for binary classification tasks. Training data is meticulously curated, consisting of samples of single and multiple objects. The labeled single-object samples are the exemplars in the training sets, and the labeled multi-object samples consist of randomly cropped patches and the entire image. 
To ensure that the class-agnostic counting is maintained, the training data is split for training and evaluation with disjoint samples, ensuring robust exemplar assessment. The classification results for positive candidate boxes $b^p \in B^p$ can be formulated as:
\begin{equation}
 \delta \left(b^p \right) = \mathrm{FFN} \left(\mathrm{Clip}\text{-}\mathrm{vit} \left(b^p \right) \right), 
\end{equation}
and the filtered set $B_\mathrm{new}$ contains bounding boxes $b^p$
that are conditioned on the classification results, which can be formulated as:
\begin{equation}
 B^p_\mathrm{new} \gets B^p_\mathrm{new} \cup \left\{b \vert \delta \left(b^p \right) = 1 \right\}, 
\end{equation}
where the symbol $\gets$ signifies the update operation for the set $B^p_\mathrm{new}$, and the set builder notation $\{b \vert \delta(b^p) = 1\}$ represents the collection of bounding boxes for which $\delta(b^p)$ predicts a positive outcome.


\subsection{Noise Suppression Module}
\label{Sec3.4}

In the context of the EEM, text-image alignment is redefined as object-image alignment by identifying positive $B^p$ and negative $B^n$ exemplars. 
We delves into generating positive and negative density maps and alleviating the noise introduced by the negative exemplars.

Initially, for each image $I_i$, we select the top three patches with the highest $S^p$ from the positive candidate boxes $B_\mathrm{new}^p$ as positive exemplars $E^p = \{b^p_i\}_{i = 1}^k$, and the top three patches with the highest $S^n$ from the negative candidate boxes $B^n_\mathrm{filtered}$ as negative exemplars $E^n = \{b^n_i\}_{i = 1}^k$. Following CounTR~\cite{liu2022countr}, we build the Counter $\Gamma(\cdot)$ with feature interaction to fuse information from both image encoders.
Specifically, we merge encoder outputs by using image features as queries and the linear projections of sample features as keys and values, ensuring dimension consistency with image features, in accordance with the self-similarity principle in counting, which can be formulated as:
\begin{equation}
	\bm{F}_\mathrm{fuse} = \Gamma_\mathrm{fuse}(\bm{F}_\mathrm{query}, \bm{W}^k \bm{F}_{\mathrm{key}}, \bm{W}^v \bm{F}_\mathrm{value}) \in \mathbb{R}^{M \times D}, 
\end{equation}
where $\bm{F}$ denotes the feature representations, $\bm{W}^k$ and $\bm{W}^v$ are the learnable weights for keys and values from $\{E^p, E^n\}$, $M$ denotes the number of tokens, $D$ is the feature dimensionality, and $\mathbb{R}^{M \times D}$ the space of the feature matrix. The decoder outputs the density heatmap after up-sampling the fused features to the input image's dimensions:
\begin{equation}
	D^n_i = \Gamma_\mathrm{decode} \left(\bm{F}_\mathrm{fuse}^n \right), \quad
	D^p_i = \Gamma_\mathrm{decode} \left(\bm{F}_\mathrm{fuse}^p \right).
\end{equation}

\textbf{Contrastive Learning and Loss Functions.}
The objective of the NSM in VA-Count is to reduce the impact of noise in images on counting performance while ensuring the accuracy of density map predictions. To achieve this, a contrastive loss $\mathcal{L}_C$ is proposed, using specified class density maps as positive samples and non-specified class density maps as negative samples. This involves maximizing the similarity between positive density maps and the ground-truth density maps and minimizing the similarity between negative density maps and the ground-truth density maps, as detailed in \cref{eq:lc}. To guide density map generation, we use the loss method from CounTR~\cite{liu2022countr}.


The density loss $\mathcal{L}_D$ is calculated as the mean squared error between each pixel of the density map $D_i^p$ generated for positive samples and the ground-truth density map $D_i^g$, as shown in \cref{eq:ld}. $H$ and $W$ respectively denote the height and width of the density map. 
\begin{equation}
\label{eq:lc}
	\mathcal{L}_C({D}^p_i, D^g_i, D^n_i) = -\log \frac{\exp \mathrm{sim} \left(D^p, D^g \right)}{\exp \mathrm{sim} \left(D^p, D^g \right) + \exp \mathrm{sim}\left(D^n, D^g \right)}, 
\end{equation}
\begin{equation}
\label{eq:ld}
	\mathcal{L}_D \left({D}^p_i, D^g_i \right) = \frac{1}{HW} \sum \left\|D^p_i - {D}^g_i \right\|^2_2, 
\end{equation}
\begin{equation}
	\mathcal{L}_\mathrm{total} \left({D}^p_i, D^g_i, D^n_i \right) = \mathcal{L}_C + \mathcal{L}_D.
\label{eq:total_loss}
\end{equation}

\section{Experimental Result}
\subsection{Datasets and Implementation Details}


\textbf{Datasets.}
\textbf{FSC-147}~\cite{hobley2022learning} dataset is tailored for class-agnostic counting with 6,135 images and 147 classes. Unique for its non-overlapping class subsets, it provides class labels and dot annotations for zero-shot counting using textual prompts. 

\textbf{CARPK}~\cite{hsieh2017drone} dataset offers a bird's-eye view of 89,777 cars in 1,448 parking lot images, testing the method's cross-dataset transferability and adaptability.


\textbf{Evaluation Metrics.}
Following previous class-agnostic object counting methods~\cite{nguyen2022few}, the evaluation metrics employed are Mean Absolute Error (MAE) and Root Mean Square Error (RMSE). 
MAE is widely used to assess model accuracy, while RMSE evaluates model robustness.

\textbf{Exemplar Enhancement Module} uses Grounding DINO\footnote{https://github.com/IDEA-Research/GroundingDINO} for bounding box proposals, setting the threshold $\tau_l$ to 0.02.
For negative sample filtering, the IoU threshold $\tau_\mathrm{iou}$ is set to 0.5. The single object classifier employs CLIP ViT-B/16\footnote{https://github.com/openai/CLIP} as its backbone, with an FFN comprising two linear layers, trained over 100 epochs at a learning rate of e{-4}. The dataset is partitioned in a 7:3 ratio.

\textbf{Noise Suppression Module} follows CounTR's~\cite{liu2022countr} two-stage training: MAE pretraining and AdamW~\cite{loshchilov2017decoupled}-optimized fine-tuning. It is trained on \textsc{FSC-147} with a learning rate of $10^{-5}$, batch size of 8, on an NVIDIA RTX L40 GPU.

\subsection{Comparison with the State-of-the-Arts}
For the performance evaluation of our method, it is benchmarked against a variety of state-of-the-art few-shot and zero-shot counting methods
on \textsc{FSC-147}. Additionally, we evaluate our method in comparison with class-specific counting models on \textsc{CARPK}.
\begin{table}[t]
	\centering
	\caption{Quantitive results of our VA-Count and other state-of-the-art competitors on \textsc{FSC-147}. The F-S, R-F, and Z-S are abbreviated for Few-shot, Reference-free, and Zero-shot settings. The best results for each scheme and the second-best results at the zero-shot setting are highlighted in bold and {underline}.}%
	\setlength{\tabcolsep}{1.2pt}
	\begin{tabular}{clcccccccc}
	\toprule
	\multirow{2}[2]{*}{Scheme} & \multirow{2}[2]{*}{Method} & \multirow{2}[2]{*}{Venue} & \multirow{2}[2]{*}{Shot} & \multicolumn{2}{c}{Val Set} & \multicolumn{2}{c}{Test Set} & \multicolumn{2}{c}{Avg} \\
	\cmidrule(lr){5-6} \cmidrule(lr){7-8} \cmidrule(lr){9-10}
	 & & & & MAE & RMSE & MAE & RMSE & MAE & RMSE \\
	\midrule
	\multirow{8}[2]{*}{F-S}
	 & FamNet~\cite{ranjan2021learning} & CVPR'21 & 3 & 24.32 & 70.94 & 22.56 & 101.54 & 23.44 & 86.24 \\
	 & CFOCNet~\cite{yang2021class} & WACV'21 & 3 & 21.19 & 61.41 & 22.10 & 112.71 & 21.65 & 87.06 \\
	 & CounTR~\cite{liu2022countr} & BMVC'22 & 3 & 13.13 & 49.83 & 11.95 & 91.23 & 12.54 & 70.53 \\
	 & LOCA~\cite{djukic2023low} & ICCV'23 & 3 & \textbf{10.24} & \textbf{32.56} & {10.97} & {56.97} & {10.61} & {44.77} \\
	 & SAM~\cite{shi2024training} & WACV'24 & 3 & - & - & 19.95 & 132.16 & 19.95 & 132.16 \\
	 & PseCo~\cite{huang2023point} & CVPR'24 & 3 & 15.31 & 68.34 & 13.05 & 112.86 & 14.18 & 90.60 \\
	 & CACViT~\cite{wang2024vision} & AAAI'24 & 3 & 10.63 & 37.95 & \textbf{9.13} & \textbf{48.96} & \textbf{9.88} & \textbf{43.46} \\
 \cmidrule{2-10}
	 & FamNet~\cite{ranjan2021learning} & CVPR'21 & 1 & 26.05 & 77.01 & 26.76 & 110.95 & 26.41 & 93.98 \\
 \midrule
 \multirow{5}{*}{R-F}
	 & FamNet~\cite{ranjan2021learning} & CVPR'21 & 0 & 32.15 & 98.75 & 32.27 & 131.46 & 32.21 & 115.11 \\
	 & RepRPN-C~\cite{ranjan2022exemplar} & ACCV'22 & 0 & 29.24 & 98.11 & 26.66 & 129.11 & 27.95 & 113.61 \\
	 & CounTR~\cite{liu2022countr} & BMVC'22 & 0 & 18.07 & 71.84 & \textbf{14.71} & 106.87 & \textbf{16.39} & 89.36 \\ 
  &  RCC~\cite{hobley2022learning} & CVPR'23 & 0 & 17.49 & 58.81 & 17.12 & 104.53 & 17.31 & 81.67 \\
	 & LOCA~\cite{djukic2023low} & ICCV'23 & 0 & \textbf{17.43} & \textbf{54.96} & 16.22 & \textbf{103.96} & 16.83 & \textbf{79.46} \\
	\midrule
	\multirow{4}{*}{Z-S} & ZSC~\cite{xu2023zero} & CVPR'23 & 0 & 26.93 & 88.63 & 22.09 & \underline{115.17} & 24.51 & 101.90 \\
	 & CLIP-Count~\cite{jiang2023clip} & MM'23 & 0 & \underline{18.79} & \textbf{61.18} & \underline{17.78} & \textbf{106.62} & 18.285 & \textbf{83.90} \\
	 & PseCo~\cite{huang2023point} & CVPR'24 & 0 & 23.90 & 100.33 & \textbf{16.58} & 129.77 & \underline{20.24} & 115.05 \\
	 & VA-Count & Ours & 0 & \textbf{17.87} & \underline{73.22} & 17.88 & 129.31 & \textbf{17.87} & \underline{101.26} \\
	\bottomrule
	\end{tabular}
	\label{tab:ExpSOTA}
\end{table}%
\begin{table}[t]
	\centering
	\caption{Quantitative results of our VA-Count and other state-of-the-art competitors on \textsc{CARPK}. $\Phi(\cdot)$ denotes the single-object classification filter. C and F denote \textsc{CARPK} and \textsc{FSC-147}, respectively.}
	\setlength{\tabcolsep}{4pt}
	\begin{tabular}{lccccccccc}
	\toprule
	\multirow{2}[2]{*}{Methods} & \multirow{2}[2]{*}{Venue} & \multirow{2}[2]{*}{Shot} & \multicolumn{2}{c}{C $\to$ C} & \multicolumn{2}{c}{F $\to$ C} \\ 
 \cmidrule(lr){4-5} \cmidrule(lr){6-7} 
 & & & MAE & RMSE & MAE & RMSE \\
	\midrule
 	FamNet~\cite{ranjan2021learning} & CVPR'21 & 3 & 18.19 & 33.66 & 28.84 & 44.47 \\
	GMN~\cite{lu2019class} & CVPR'21 & 3 & 7.48 & 9.90 & - & - \\
	BMNet+~\cite{shi2022represent} & CVPR'22 & 3 & 5.76 & 7.83 & \textbf{10.44} & \textbf{13.77} \\
	CounTR~\cite{liu2022countr} & BMVC'22 & 3 & \textbf{5.75} & \textbf{7.45} & - & - \\
	\midrule

 RCC~\cite{hobley2022learning} & CVPR'23 & 0 & 9.21 & 11.33 & 21.38 & 26.61 \\
	CLIP-Count~\cite{jiang2023clip} & MM'23 & 0 & - & - & 11.96 & 16.61 \\
		Grounding DINO~\cite{Liu2023DINO} & arXiv'24 & 0 & 29.72 & 31.60 & 29.72 & 31.60 \\
	Grounding DINO + $\Phi(\cdot)$ & Ours & 0 & 18.54 & 21.71 & 18.54 & 21.71 \\
	VA-Count & Ours & 0 & \textbf{8.75} & \textbf{10.30} & \textbf{10.63} & \textbf{13.20} \\
 \bottomrule
	\end{tabular}
	\label{tab2}
\end{table}
\begin{table}[t]
	\centering
 	\caption{Ablation study on each component’s contribution to the final results on \textsc{FSC-147}. We demonstrate the effectiveness of two parts of our framework and two types of loss: $G(\cdot)$ for Grounding DINO, $\Phi(\cdot)$ for the single-object filtering section, the density loss $\mathcal{L}_D$, and the contrastive loss $\mathcal{L}_C$.} 
	\setlength{\tabcolsep}{8pt}
	\begin{tabular}{cccccccc}
	\toprule
	\multirow{2}[2]{*}{$G(\cdot)$} & \multirow{2}[2]{*}{$\phi(\cdot)$} & \multirow{2}[2]{*}{$\mathcal{L}_D$} & \multirow{2}[2]{*}{$\mathcal{L}_{C}$} & \multicolumn{2}{c}{Val Set} & \multicolumn{2}{c}{Test Set} \\
 	\cmidrule(lr){5-6} \cmidrule(lr){7-8}
	 & & & & MAE & RMSE & MAE & RMSE \\
	\midrule
	\CIRCLE & \Circle & \Circle & \Circle & 52.82 & 134.49 & 54.48 & 159.30 \\
	\CIRCLE & \CIRCLE & \Circle & \Circle & 52.12 & 135.29 & 54.27 & 159.76 \\
	\CIRCLE & \CIRCLE & \CIRCLE & \Circle & 19.63 & 73.94 & 18.93 & \textbf{116.65} \\
	\CIRCLE & \CIRCLE & \CIRCLE & \CIRCLE & \textbf{17.87} & \textbf{73.22} & \textbf{17.88} & 129.31 \\
	\bottomrule
	\end{tabular}
	\label{tab:abalation_factors}
\end{table} %
\textbf{Quantitative Result on FSC-147.}
We evaluate the effectiveness of VA-Count on \textsc{FSC-147}, comparing it with state-of-the-art counting methods as detailed in \cref{tab:ExpSOTA}. Our method surpasses the exemplar-discovery method ZSC~\cite{xu2023zero}, demonstrating that the exemplars found by VA-Count are of higher quality. VA-Count achieves the best performance in MAE and second in RMSE, validating our method's effectiveness. Despite being second in RMSE, it still outperforms ZSC. In comparison with CLIP-Count~\cite{jiang2023clip}, VA-Count, due to some noise introduction, has a few inferior samples but, overall, surpasses CLIP-Count in performance.
\textbf{Quantitative Result on CARPK.}
In \cref{tab2}, VA-Count's cross-domain and non-cross-domain performance on \textsc{CARPK} are compared with previous methods. In the zero-shot group, VA-Count achieves the best performance, particularly with its cross-domain performance methoding that of the few-shot group, demonstrating its outstanding transferability. It is worth noting that employing $\Phi(\cdot)$ significantly reduces errors compared to directly using the Grounding DINO~\cite{Liu2023DINO} method. In the absence of any training data, VA-Count outperforms FamNet~\cite{ranjan2021learning} in the cross-domain group.


\textbf{Ablation Study.}
We conduct both quantitative and qualitative analyses on the contributions of each component in our proposed VA-Count, which includes the Grounding-DINO candidate box extraction and filtering module. The quantitative outcomes are presented in \cref{tab:abalation_factors}.
Using only Grounding DINO method (first row) achieves an error of 52.82 without training, which, although not as accurate as regression-based methods, ensures the detection of relevant objects. Performance improves slightly after adding a single-object classification filter (second row). With training based on $\mathcal{L}_D$, it already meets counting requirements.
In \cref{tab2}, we compare using Grounding DINO alone and with a single-object classification filter on \textsc{CARPK} (last three rows). Our binary classifier significantly improves performance, reducing MAE and RMSE by about 10.


\begin{figure}[t]
	\centering
	\includegraphics[width = 0.9\textwidth]{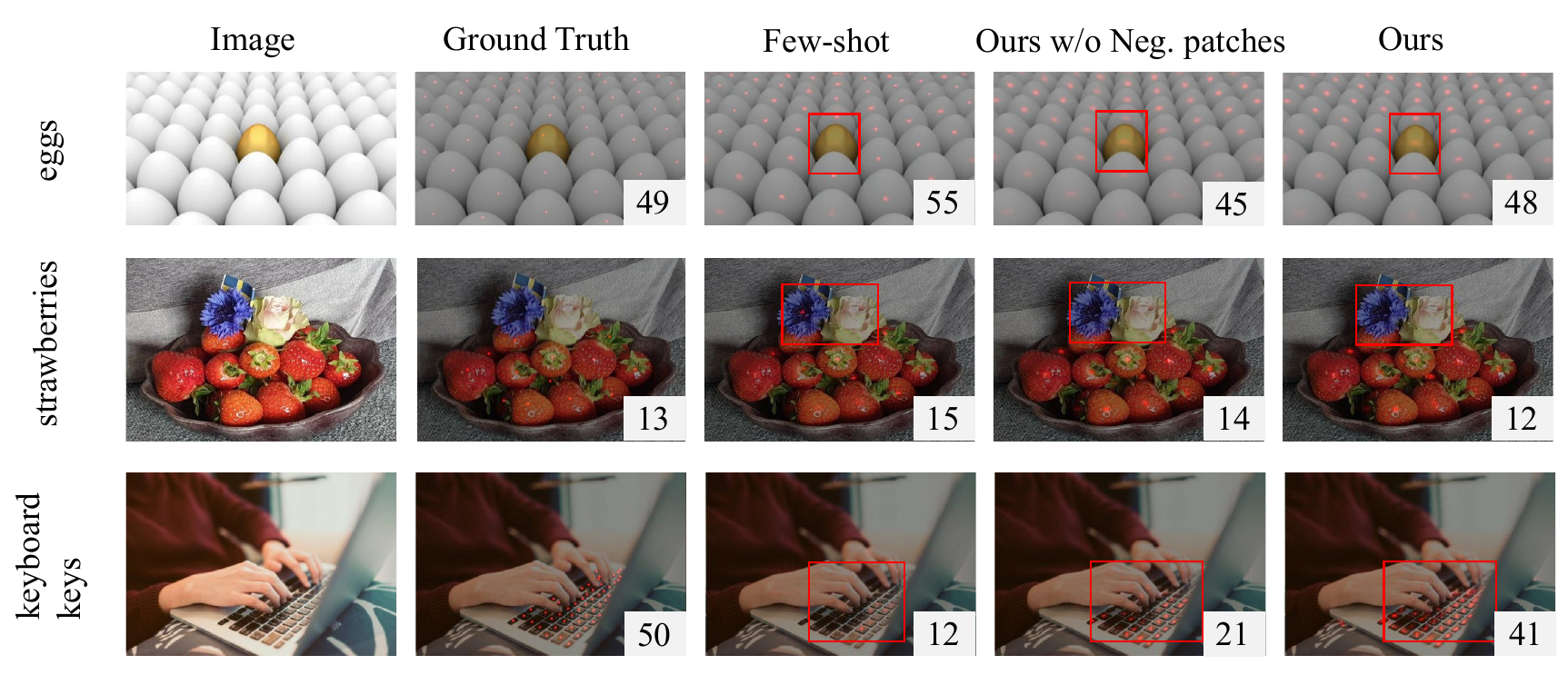}
	\caption{Illustration of heatmaps compared with few-shot method~\cite{liu2022countr} on \textsc{FSC-147}. The predicted density map is overlaid on the original RGB image. (Best viewed in zoom in)}
	\label{fig:visualization_heatmap}
	\label{tab:ab}
\end{figure}
\begin{figure}[t]
	\centering
	\includegraphics[width = 0.86\textwidth]{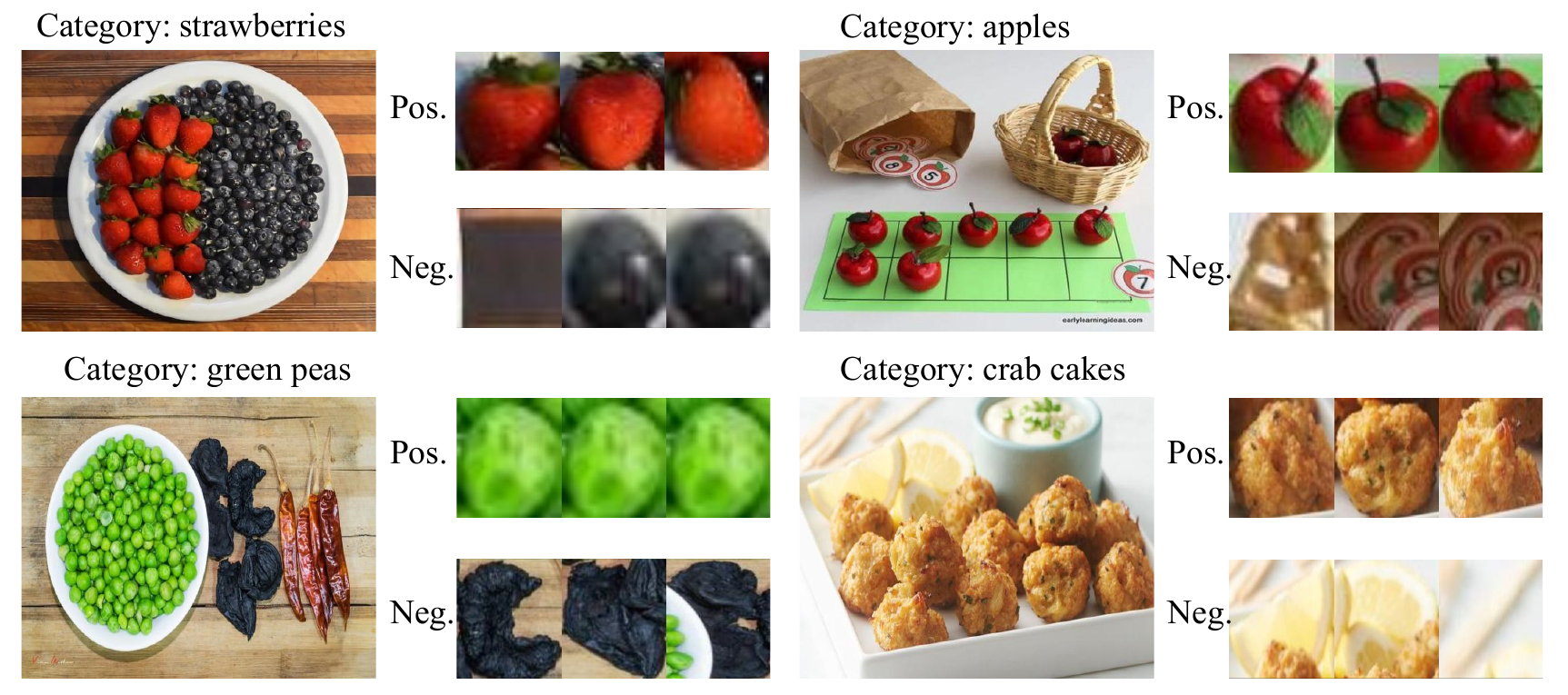}
	\caption{Illustration of the final positive (Pos.) and negative (Neg.) exemplars for images on \textsc{FSC-147}.}
	\label{fig:visualization_patch}
\end{figure}

\subsection{Qualitative Analysis}
\textbf{Analysis of the zero-shot performance.}
To further ensure the effectiveness of the proposed VA-Count framework, we visualize qualitative results in \cref{{tab:ab}}. We provide a side-by-side comparison of the proposed VA-Count against the few-shot counting method~\cite{liu2022countr}. VA-Count achieves a remarkable resemblance to the ground truth, showcasing the method's nuanced understanding of object boundaries and densities and being less affected by the background noise.
Specifically, the first row shows there exists a golden egg drowned by white eggs. The few-shot method struggled with this nuanced differentiation, failing to recognize the golden egg distinctly. 
In the second row, strawberries near flowers also confound the few-shot method. These examples emphasize VA-Count's superior ability to identify and differentiate between objects with minor differences.
The third row presents a challenging scenario with dense keys partially occluded by hands. This situation tests the model's ability to count tiny, closely situated objects under partial occlusion, showcasing VA-Count's advanced capability to accurately identify and count such challenging objects, which is significantly better than the few-shot method.
These results underscore the impact of the exemplar selection and incorporate negative patches into VA-Count, significantly refining the model's object counting and localization capabilities, and highlighting the innovation of VA-Count to zero-shot object counting.

\begin{figure}[t]
	\centering
	\includegraphics[width = 0.9\textwidth]{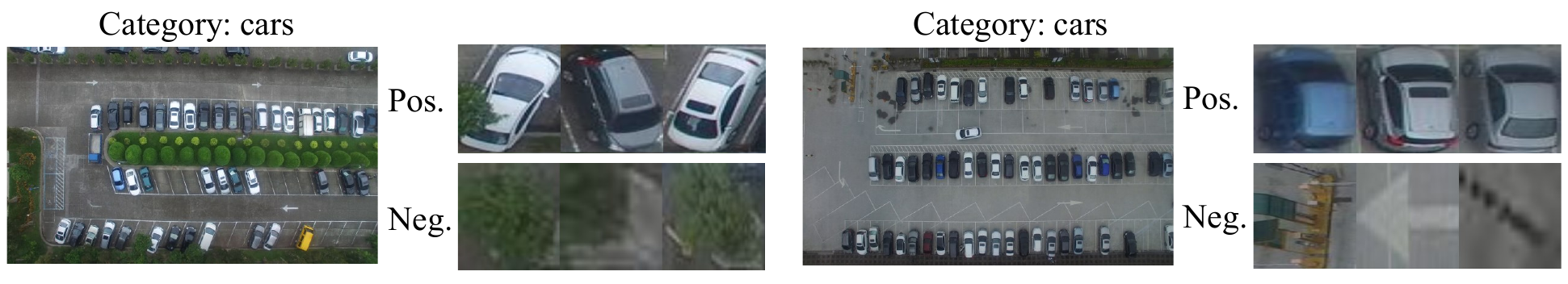}
	\caption{Illustration of the final positive (Pos.) and negative (Neg.) exemplars for images on \textsc{CARPK}.}
	\label{fig:visualization_patch_car_row}
\end{figure}

\textbf{Analysis of Positive and Negative Exemplars.}
To make our experiment more straightforward, we also conduct a qualitative analysis of the patch selection. As shown in \cref{fig:visualization_patch} and \cref{fig:visualization_patch_car_row}, we illustrate selected positive and negative patches for various categories under a zero-shot setting.
Taking a closer look at the positive patches for categories such as crab cakes and green peas, the results show a high degree of accuracy in the model's ability to isolate and highlight the regions containing the target objects. This precision underscores the effectiveness of VA-Count framework in discerning relevant features amidst complex backgrounds, affirming its robustness in the exemplar discovery.
Negative patches, especially from categories like strawberries and crab cakes, highlight the model's challenges with visually similar or overlapping areas not in the target category, underscoring the need for improved discriminative abilities. 
This analysis underscores our paper's impact on zero-shot object counting and the importance of refining visual learning and exemplar selection for future advancements.


\begin{figure}[t]
	\centering
	\includegraphics[width = 0.9\textwidth]{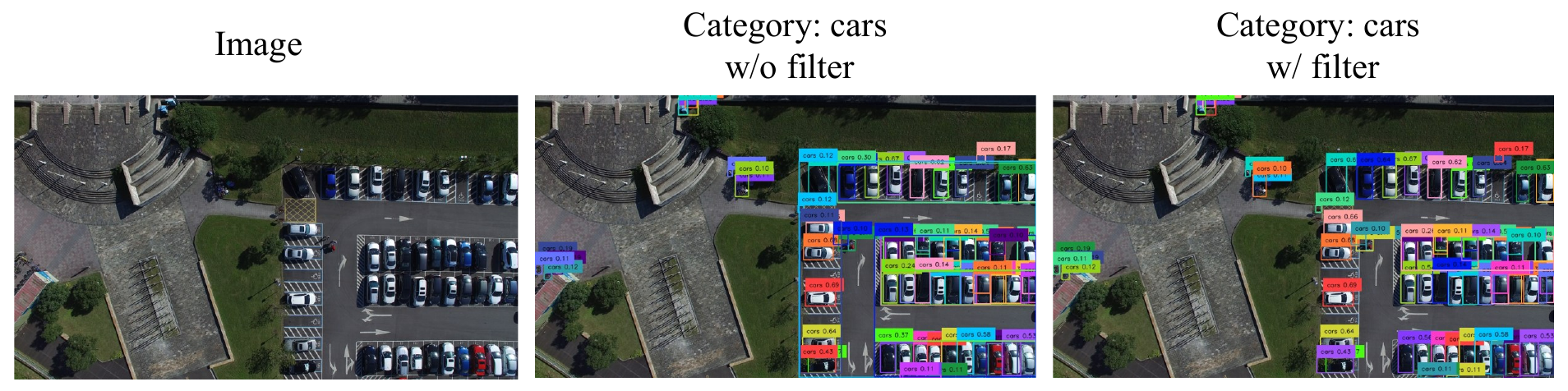}
	\vspace{-8pt}
	\caption{Illustration of the comparison of the candidate boxes before and after single object exemplar filter on \textsc{CARPK}.}
	\label{fig:visualize_filter}
\end{figure}

\textbf{Effective of the object exemplar filter.}
The effectiveness of the object exemplar filter is further evaluated by comparing visualization grounding results with and without the filter. \cref{fig:visualize_filter} illustrates this comparison for the category of cars on \textsc{CARPK}. Images without the filter show multiple cars within a single bounding box, indicating Grounding DINO's~\cite{Liu2023DINO} inability to isolate individual objects effectively. Conversely, images with the filter applied demonstrate a significant improvement, with bounding boxes accurately encompassing single cars. This clear distinction highlights the binary classifier's crucial role in ensuring precise object counting by enforcing the single-object criterion within each exemplar, validating the filter's contribution to enhancing the model's accuracy and reliability in VA-Count framework.

\section{Conclusion}
This paper addresses the challenges in class-agnostic object counting by introducing the Visual Association-based Zero-shot Object Counting (VA-Count) framework. VA-Count effectively balances the need for scalability across arbitrary classes with the establishment of robust visual connections, overcoming the limitations of existing Zero-shot Object Counting (ZOC) methods. VA-Count comprises an Exemplar Enhancement Module (EEM) and a Noise Suppression Module (NSM), which are dedicated to refining exemplar identification and mitigating adverse impacts, respectively. The EEM utilizes advanced Vision-Language Pretaining models like Grounding DINO for scalable exemplar discovery, while the NSM mitigates the impact of erroneous exemplars through contrastive learning.
VA-Count shows promise in zero-shot counting, performing well on three datasets and offering precise visual associations and scalability. 
In the future, we will explore and better utilize advanced visual language models.



\section*{Acknowledgments}
This work was supported in part by the National Natural Science Foundation of China under Grant 62271361, the Sanya Yazhou Bay Science and Technology City Administration scientific research project under Grant 2022KF0021, the Guangdong Natural Science Funds for Distinguished Young Scholar under Grant 2023B1515020097, and the National Research Foundation Singapore under the AI Singapore Programme under Grant AISG3-GV-2023-011.



%
%

\newpage
\bibliographystyle{splncs04}
\bibliography{counting}
\newpage
\appendix
\section{Appendix}
\subsection{Overview}
\begin{itemize}
	\item Analysis of Density Maps (\cf \cref{sec1})
 	\item Analysis of Negative Sample Density Maps (\cf \cref{sec2})
 	\item Analysis of Positive and Negative Samples (\cf \cref{sec3})
 	\item Ablation Study on IoU Threshold (\cf \cref{sec4})
 	\item Ablation Study on Thresholds for Grounding DINO (\cf \cref{sec5})
 	\item Transfer experiments on crowd datasets (\cf \cref{sec6})
 	\item Limitation (\cf \cref{sec7})
\end{itemize}

\begin{figure}[t]
	\centering
	\includegraphics[width = \textwidth]{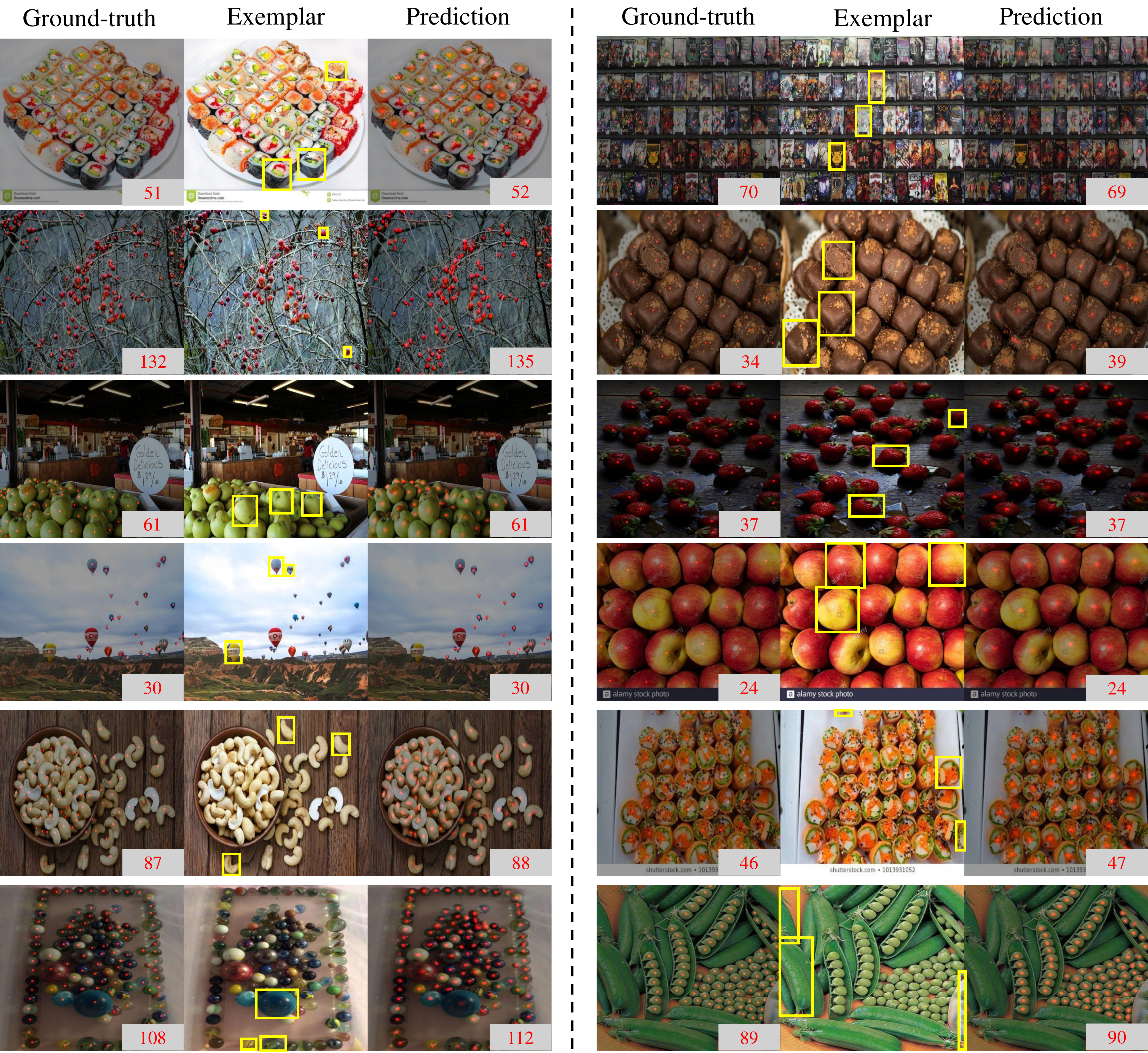}
	\caption{Illustration of the found exemplars for images on \textsc{FSC-147}, along with the density maps.}
	\label{fig:vis_density}
\end{figure}

\subsection{Analysis of Density Maps}
\label{sec1}
\cref{fig:vis_density} demonstrates the efficacy of VA-Count in generating density maps, where it is evident that our method yields estimations closely aligned with ground-truth densities across a spectrum of scenarios: handling of irregularly shaped objects (first and fifth rows), navigation through complex environmental backgrounds (images two, three, and four from the left), and accurate depiction of densely clustered objects (images two, three, and four from the right). The exemplars utilized are of exceptional quality. Notably, even in scenarios with significant object scale variability, as depicted in the lower left image, the algorithm successfully approximates true density values. Moreover, the robustness of VA-Count is highlighted in the rightmost sixth image, where despite the selection of exemplars with minor inaccuracies, the density map produced is of high fidelity. 
This demonstrates VA-Count's ability to maintain the intrinsic correlation between exemplars and original images, ensuring minor selection errors in exemplars have minimal impact on density estimation accuracy.

\begin{figure}[t]
	\centering
	\includegraphics[width = \textwidth]{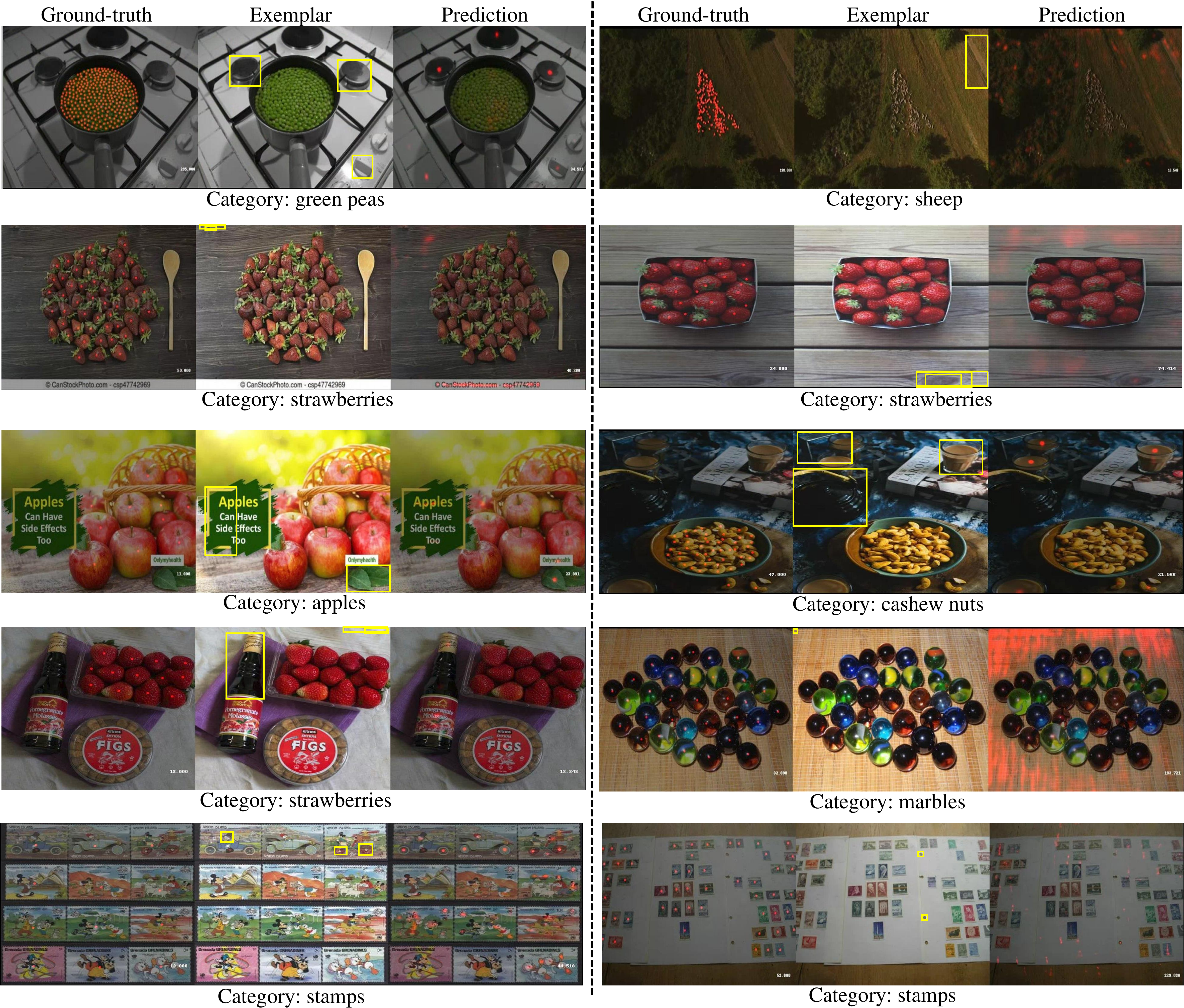}
	\caption{Illustration of the final negative exemplars for images on \textsc{FSC-147}, along with the density maps.}
	\label{fig:vis}
\end{figure}

\subsection{Analysis of Negative Sample Density Maps}
\label{sec2}
\cref{fig:vis} shows the negative exemplar and the corresponding density map display. The figure demonstrates that when the exemplar is not a sample of the corresponding category, it will not find the specified category, but instead will locate the area corresponding to the negative exemplar and generate a density map. 
When objects belonging to different categories are present within an image (as observed in positions left 1, left 4, left 5, and right 3), density maps specific to those categories are produced. Conversely, in scenarios devoid of distinguishable objects, where only the background is visible, the generated density maps correlate directly with the designated regions.

\begin{figure}[t]
	\centering
	\includegraphics[width = \textwidth]{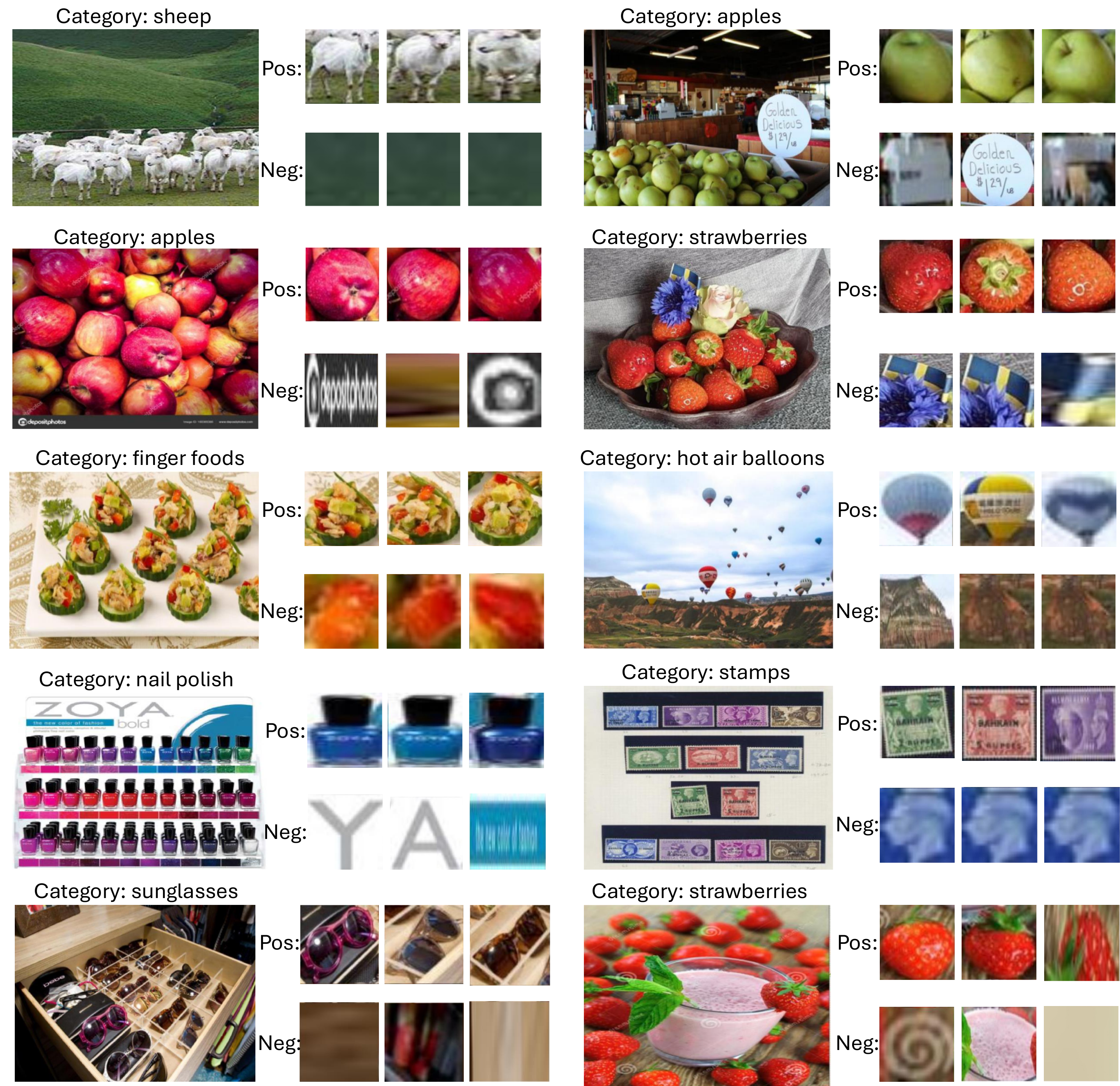}
 	\caption{Illustration of the positive (Pos.) and negative (Neg.) exemplars for images on \textsc{FSC-147}.}
	\label{fig:vis2}
\end{figure}

\subsection{Analysis of Positive and Negative Samples}
\label{sec3}
\cref{fig:vis2} illustrates the selection process for positive and negative samples. From the figure, it is evident that our method identifies positive samples as individual objects of the specified category, performing well not only for regular objects but also for items like nail polish, sunglasses, and stamps. In selecting negative samples, when objects of other categories are present in the image, our method can identify these objects as negative samples (as seen in left 2, left 3, right 2, right 3, and right 4). This demonstrates that VA-Count not only selects high-quality positive exemplars but also effectively avoids positive samples while selecting potentially confusing objects as negative samples.

\begin{table}[t]
	\centering
 	\caption{Ablation study on the contribution of the IoU threshold $\tau_\mathrm{iou}$ for negative sample selection to the final results on \textsc{FSC-147}. We present the MAE and RMSE across the validation and test sets for thresholds ranging from 0.1 to 0.9, as well as their average performance. The best results are highlighted in \textbf{bold}, and the second-best are \underline{underlined}.} 
	\setlength{\tabcolsep}{10pt}
	\begin{tabular}{ccccccc}
	\toprule
	\multirow{2}[2]{*}{$\tau_\mathrm{iou}$} & 
 \multicolumn{2}{c}{Val Set} & 
 \multicolumn{2}{c}{Test Set} & 
 \multicolumn{2}{c}{Avg} \\
 	\cmidrule(lr){2-3} \cmidrule(lr){4-5}
 \cmidrule(lr){6-7}
	 & MAE & RMSE & MAE & RMSE & MAE & RMSE \\
	\midrule
	0.1 & 18.83 & \textbf{72.26} & 20.27 & 130.39 & 19.55 & 101.33 \\
	0.2 & 18.56 & 77.01 & 18.73 & \underline{125.83} & 18.64 & 101.42 \\
	0.3 & 19.89 & 77.23 & 18.52 & \textbf{125.41} & 19.20 & 101.32 \\
 0.4 & \underline{18.26} & 75.61 & \textbf{17.54} & 127.47 & \underline{17.90} & 101.54 \\
	0.5 & \textbf{17.87} & \underline{73.22} & \underline{17.88} & 129.31 & \textbf{17.87} & \textbf{101.26} \\
 0.6 & 18.55 & 73.90 & 19.10 & 129.32 & 18.82 & 101.61 \\
 0.7 & 18.97 & 74.91 & 18.31 & 128.78 & 18.64 & 101.85 \\
 0.8 & 21.28 & 74.51 & 20.52 & 128.00 & 20.90 & \textbf{101.26} \\
 0.9 & 22.30 & 74.48 & 20.96 & 128.31 & 21.63 & 101.40 \\
	\bottomrule
	\end{tabular}
	\label{tab:abalation_1}
\end{table} %

\subsection{Ablation Study on IoU Threshold}
\label{sec4}
The Intersection over Union (IoU) threshold plays a critical role in determining the quality of negative sample selection. 
\cref{tab:abalation_1} illustrates the influence of varying IoU thresholds on the accuracy of object counting, presenting data for the Mean Absolute Error (MAE) and Root Mean Square Error (RMSE) across both the validation and test datasets. Notably, the MAE demonstrates a non-linear trend, initially rising before diminishing, with the optimal performance observed at an IoU threshold of 0.5. 
In contrast, the RMSE experiences fluctuations, attributable to the varying quality of density maps influenced by the selection of negative samples. Such variations in density map quality introduce a stochastic element to the errors, thereby causing the observed fluctuations in RMSE.

\begin{table}[t]
	\centering
 	\caption{Ablation study on the contribution of the grounding DINO threshold for sample selection to the final results on \textsc{FSC-147}. We present the MAE and RMSE across the validation and test sets for Logits thresholds $\tau_l$ ranging from 0.01 to 0.05, as well as their average performance. The best results are highlighted in \textbf{bold}, and the second-best are \underline{underlined}.} 
	\setlength{\tabcolsep}{10pt}
	\begin{tabular}{ccccccc}
	\toprule
	\multirow{2}[2]{*}{$\tau_l$} & 
	\multicolumn{2}{c}{Val Set} & \multicolumn{2}{c}{Test Set} & \multicolumn{2}{c}{Avg} \\
 	\cmidrule(lr){2-3} \cmidrule(lr){4-5}
	\cmidrule(lr){6-7}
	 & MAE & RMSE & MAE & RMSE & MAE & RMSE \\
	\midrule
	0.01 & 27.36 & 76.41 & 27.10 & 129.98 & 27.23 & 103.20 \\
	0.02 & \textbf{17.87} & \textbf{73.22} & \textbf{17.88} & \underline{129.31} & \textbf{17.87} & \textbf{101.26} \\
	0.03 & \underline{19.74} & 77.06 & \underline{18.25} & 129.77 & \underline{18.99} & 103.42 \\
	0.04 & 22.84 & \underline{76.26} & 20.26 & \textbf{128.69} & 21.55 & \underline{102.48} \\
 0.05 & 25.60 & 86.45 & 21.25 & 130.79 & 23.43 & 108.62 \\
	\bottomrule
	\end{tabular}
	\label{tab:abalation_2}
\end{table} %

\subsection{Ablation Study on Thresholds for Grounding DINO}
\label{sec5}

The selection of logits thresholds for Grounding DINO is identified as a pivotal factor in curating exemplars. Excessively high thresholds hinder the selection of samples for more challenging categories, while excessively low thresholds not only escalate computational demands but also result in an abundance of superfluous samples. To address this, we conducted the experiments detailed in \cref{tab:abalation_2}. At a threshold of 0.01, the inclusion of suboptimal exemplars significantly elevates the RMSE. Conversely, setting the threshold at 0.05 leads to a considerable overall error, as it precludes the selection of category-specific exemplars in certain images. The thresholds of 0.02, 0.03, and 0.04 exhibit comparatively lower MAE and RMSE values, with the optimal error minimization achieved at a threshold of 0.02. This nuanced method underscores the importance of a balanced threshold setting in enhancing the efficacy of exemplar selection within the Grounding DINO framework.

\begin{table}[t]
	\centering
	\caption{Transfer experiments on crowd datasets. FSC, SHA, and SHB denote \textsc{FSC-147} and \textsc{ShanghaiTech A} and \textsc{ShanghaiTech B}, respectively.}
	\setlength{\tabcolsep}{10pt}
	\begin{tabular}{lcccc}
	\toprule
	\multirow{2}[2]{*}{Method} & \multicolumn{2}{c}{FSC $\to$ SHA} & \multicolumn{2}{c}{FSC $\to$ SHB} \\
	\cmidrule(lr){2-3} \cmidrule(lr){4-5} 
	& MAE & RMSE & MAE & RMSE \\
	\midrule
	{RCC~\cite{hobley2022learning}} & 240.1 & 366.9 & 66.6 & 104.8 \\
	{VA-Count(Ours)} & \textbf{213.0} & \textbf{360.8} & \textbf{40.3} & \textbf{68.1} \\
	\bottomrule
	\end{tabular}
	\label{tab:abalation_3}
\end{table}

\subsection{Transfer experiments on crowd datasets}
\label{sec6}
To evaluate VA-Count's transferability, \cref{tab:abalation_3} presents the transfer experiments from the FSC dataset to \textsc{ShanghaiTech} crowd dataset. Our method achieved competitive results without any fine-tuning.

\begin{figure}[t]
	\centering
	\includegraphics[width = \textwidth]{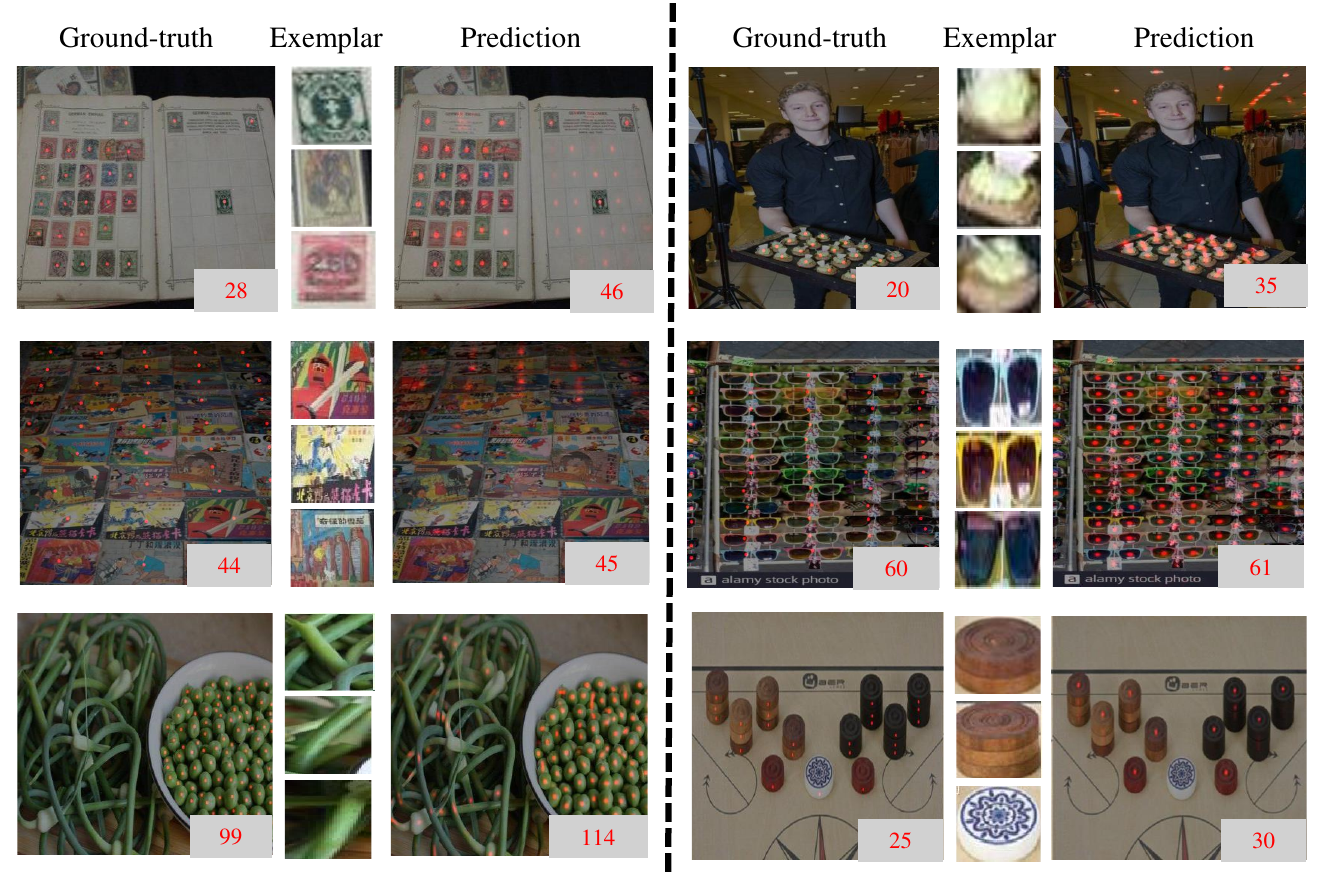}
	\caption{Illustration of the error density map on \textsc{FSC-147}.}
	\label{fig:vis4}
\end{figure} 

\subsection{Limitation}
\label{sec7}

To delve into the limitations of VA-Count, \cref{fig:vis4} showcases images with notable inaccuracies, highlighting three primary constraints in the algorithm's efficacy. Firstly, there is the challenge of background noise. Despite the strategic use of negative samples to mitigate errors from non-object classes, the algorithm remains excessively responsive to clear objects (first row). Secondly, the issue of density map numerical uncertainty is evident. As illustrated in the second row, despite both images having a mere count error of 1, the quality of their density maps is suboptimal. Specifically, the left image poorly locates a larger object in the foreground, while the right image incorrectly identifies two points of focus for a single pair of sunglasses, diverging from the ground-truth which associates one focal point per pair of sunglasses. Lastly, exemplar inaccuracies persist. While our method achieves exemplar identification quality on par with annotated bounding boxes in most images, some discrepancies remain. For instance, as depicted on the left, entire strings of peas are mistakenly identified as exemplars, and on the right, stacked items, not individual objects due to their blurred edges, are erroneously treated as singular targets. These limitations represent key areas for our ongoing and future refinement efforts.


\end{document}